\documentclass{article} 
\usepackage{colm2024_conference}

\usepackage{microtype}
\usepackage{hyperref}
\usepackage{url}
\usepackage{booktabs}
\definecolor{darkblue}{rgb}{0, 0, 0.5}
\hypersetup{colorlinks=true, citecolor=darkblue, linkcolor=darkblue, urlcolor=darkblue}

\newcommand{\eg}{\textit{e.g.}}

\newcommand{\methodname}{ECHO}

\usepackage{colortbl}
\usepackage{graphicx}
\usepackage{enumitem}
\usepackage{multirow}
\usepackage[linesnumbered,ruled,vlined]{algorithm2e}

\renewcommand{\cite}{\citep}

\definecolor{mygray}{RGB}{226, 226, 226}
\definecolor{myred}{RGB}{252, 142, 142}
\definecolor{mygreen}{RGB}{147, 255, 143}
\definecolor{myblue}{RGB}{144, 155, 255}
\definecolor{myyellow}{RGB}{253, 253, 143}
\definecolor{mypurple}{RGB}{255, 142, 250}

\usepackage{pgfplots}
\usepgfplotslibrary{polar}
\usepackage{subcaption}
\pgfplotsset{compat=newest}
\SetKwComment{Comment}{$\triangleright$\ }{}

\title{How Well Can LLMs Echo Us? \\ \underline{E}valuating AI \underline{Ch}atbots' R\underline{o}le-Play Ability with \underline{ECHO}}

\author{Man Tik Ng\footnotemark[1], Hui Tung Tse\footnotemark[1], Jen-tse Huang\footnotemark[2], Jingjing Li, Wenxuan Wang, Michael R. Lyu \\
Department of Computer Science and Engineering, The Chinese University of Hong Kong\\
\texttt{\{derek331, httse0930, lijj\}@link.cuhk.edu.hk} \quad \texttt{\{jthuang, wxwang, lyu\}@cse.cuhk.edu.hk} \\
}

\footnotetext[1]{Equal contribution.}
\footnotetext[2]{Corresponding author.}

\colmfinalcopy 
\begin{document}

\maketitle

\begin{abstract}
The role-play ability of Large Language Models (LLMs) has emerged as a popular research direction.
However, existing studies focus on imitating well-known public figures or fictional characters, overlooking the potential for simulating ordinary individuals.
Such an oversight limits the potential for advancements in digital human clones and non-player characters in video games.
To bridge this gap, we introduce {\methodname}, an evaluative framework inspired by the Turing test.
This framework engages the acquaintances of the target individuals to distinguish between human and machine-generated responses.
Notably, our framework focuses on emulating average individuals rather than historical or fictional figures, presenting a unique advantage to apply the Turing Test.
We evaluated three role-playing LLMs using {\methodname}, with GPT-3.5 and GPT-4 serving as foundational models, alongside the online application GPTs from OpenAI.
Our results demonstrate that GPT-4 more effectively deceives human evaluators, and GPTs achieves a leading success rate of 48.3\%.
Furthermore, we investigated whether LLMs could discern between human-generated and machine-generated texts.
While GPT-4 can identify differences, it could not determine which texts were human-produced.
Our code and results of reproducing the role-playing LLMs are made publicly available via \url{https://github.com/CUHK-ARISE/ECHO}.
\end{abstract}

\section{Introduction}

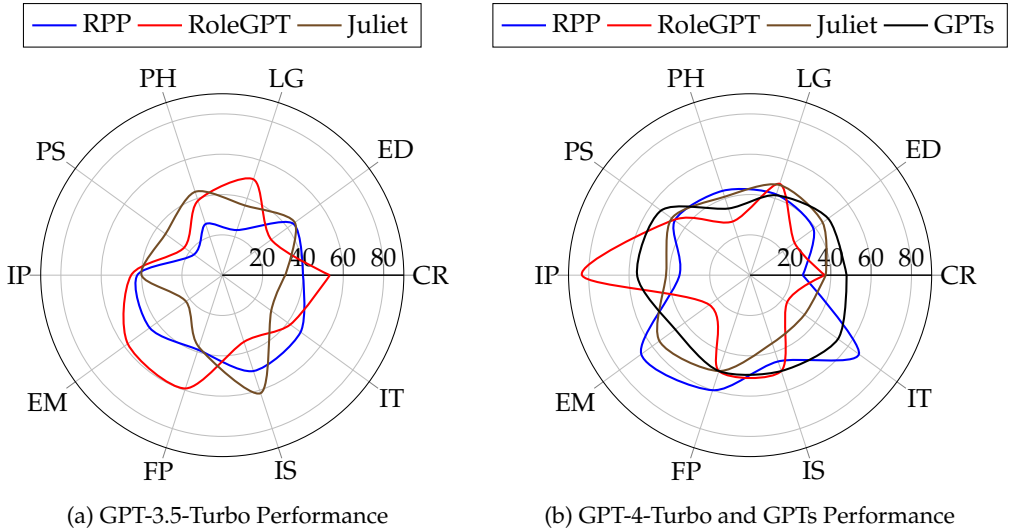
\begin{figure}[t]
    \centering
    \resizebox{1.0\textwidth}{!}{
    \begin{subfigure}{.5\textwidth}
        \centering
        \begin{tikzpicture}
            \begin{polaraxis}[
                width=6.5cm,
                height=6.5cm,
                legend style={at={(0.5,1.25)},anchor=north,legend columns=-1},
                ymax=90,
                ymin=0,
                ytick={20, 40, 60, 80},
                xtick={0,36,72,108,144,180,216,252,288,324},
                xticklabels={CR,ED,LG,PH,PS,IP,EM,FP,IS,IT},
            ]
            \addplot+[mark=none,smooth,thick] coordinates {
                (0,40)
                (36,43.48)
                (72,23.53)
                (108,26.67)
                (144,17.39)
                (180,42.11)
                (216,44.44)
                (252,38.89)
                (288,50)
                (324,48)
                (360,40)
            };
            \addlegendentry{RPP}
            \addplot+[mark=none,smooth,thick] coordinates {
                (0,53.33)
                (36,30)
                (72,50)
                (108,38.89)
                (144,23.33)
                (180,45.16)
                (216,57.89)
                (252,59.09)
                (288,34.78)
                (324,41.67)
                (360,53.33)
            };
            \addlegendentry{RoleGPT}
            \addplot+[mark=none,smooth,thick] coordinates {
                (0,31.25)
                (36,44.44)
                (72,36.36)
                (108,43.48)
                (144,34.78)
                (180,40)
                (216,22.22)
                (252,37.50)
                (288,61.54)
                (324,30)
                (360,31.25)
            };
            \addlegendentry{Juliet}
            \end{polaraxis}
        \end{tikzpicture}
        \caption{GPT-3.5-Turbo Performance}
        \label{fig:radar-3.5}
    \end{subfigure}
    
    \begin{subfigure}{.5\textwidth}
        \centering
        \begin{tikzpicture}
            \begin{polaraxis}[
                width=6.5cm,
                height=6.5cm,
                legend style={at={(0.5,1.25)},anchor=north,legend columns=-1},
                ymax=90,
                ymin=0,
                ytick={20, 40, 60, 80},
                xtick={0,36,72,108,144,180,216,252,288,324},
                xticklabels={CR,ED,LG,PH,PS,IP,EM,FP,IS,IT},
            ]
            \addplot+[mark=none,smooth,thick] coordinates {
                (0,26.09)
                (36,38.89)
                (72,42.11)
                (108,44.00)
                (144,46.15)
                (180,35.00)
                (216,66.67)
                (252,60.00)
                (288,45.00)
                (324,66.67)
                (360,26.09)
            };
            \addlegendentry{RPP}
            \addplot+[mark=none,smooth,thick] coordinates {
                (0,37.04)
                (36,27.27)
                (72,47.62)
                (108,28.00)
                (144,46.67)
                (180,83.33)
                (216,25.00)
                (252,50.00)
                (288,50.00)
                (324,22.73)
                (360,37.04)
            };
            \addlegendentry{RoleGPT}
            \addplot+[mark=none,smooth,thick] coordinates {
                (0,37.50)
                (36,44.44)
                (72,47.06)
                (108,40.91)
                (144,48.00)
                (180,41.67)
                (216,55.56)
                (252,50.00)
                (288,35.48)
                (324,33.33)
                (360,37.50)
            };
            \addlegendentry{Juliet}
            \addplot+[mark=none,smooth,thick] coordinates {
                (0,47.83)
                (36,47.83)
                (72,41.67)
                (108,34.78)
                (144,54.55)
                (180,56.00)
                (216,45.83)
                (252,50.00)
                (288,50.00)
                (324,53.85)
                (360,47.83)
            };
            \addlegendentry{GPTs}
            \end{polaraxis}
        \end{tikzpicture}
        \caption{GPT-4-Turbo and GPTs Performance}
        \label{fig:radar-4}
    \end{subfigure}
    }
    \caption{Success rates of role-playing LLMs in deceiving human evaluators. The human evaluators are instructed to identify \textbf{human}-generated responses.}
    \label{fig:radar}
\end{figure}

Large Language Models (LLMs) have recently made significant breakthroughs in the field of Artificial Intelligence (AI).
Notably, ChatGPT\footnote{\url{https://chat.openai.com/}}, one of the leading commercial models, has showcased its capabilities across different Natural Language Processing (NLP) tasks, such as information retrieval~\cite{zhu2023large}, computer programming~\cite{surameery2023use}, grammar checking~\cite{wu2023chatgpt}, and sentence translation~\cite{jiao2023chatgpt}.
Trained on extensive datasets, LLMs also demonstrate applicability beyond NLP tasks, extending to domains such as healthcare~\cite{johnson2023assessing}, education~\cite{baidoo2023education}, legal service~\cite{guha2024legalbench}, and product design~\cite{lanzi2023chatgpt}.

Given LLMs' extensive capabilities, researchers have explored their human-like abilities~\cite{huang2024humanity, huang2023emotionally} and their performance on complex tasks~\cite{wan2024b, huang2024far}.
\textit{Role-playing}, the act of changing one's behavior to fulfill a specific role, has been employed as a scenario to evaluate LLMs~\cite{shanahan2023role, wang2023incharacter} since it is a complicated task requiring various abilities.
However, the evaluation of LLMs' role-playing ability remains relatively unexplored.
Previous studies~\cite{shao2023character, wang2023rolellm} mainly focus on instructing LLMs to impersonate celebrities or fictional characters whose data are likely to be included in the training corpus of the LLMs.
As a result, the ability of LLMs to role-play as typical individuals is not well understood, limiting our evaluation of their role-playing potential.
This oversight could restrict the scope of assessing LLMs' role-playing capabilities and overlooking situations where LLMs could act as digital clones of humans, non-player characters in video games and metaverse, or, more concerningly, be used maliciously to impersonate individuals, spreading false information or damaging reputations.

Addressing this gap, our study directs LLMs to emulate real, ordinary individuals instead of famous figures, leveraging the \textit{Turing test}.
As initially proposed by \citet{turing1950computing}, this test gauges whether a machine can demonstrate intelligence indistinguishable from that of a human.
In our study, we create a role-playing LLM using the profile of an actual person and invite acquaintances of this person to discern between responses from the actual individual and the LLM.
Utilizing real-person data makes it possible to apply the Turing test and makes it easier to recruit annotators, which is advantageous over using profiles of well-known figures due to the accessibility of their acquaintances.
However, a limitation arises in multi-round dialogues, where human evaluators can easily differentiate between LLMs and actual people by posing questions LLMs cannot answer, such as queries about the current time.
This issue can shift evaluators' focus from assessing the LLMs' ability to think and act like the intended emulation target.
To address this problem, we introduce a novel framework, {\methodname}, designed to specifically evaluate LLMs' proficiency in replicating a human's thought process within a particular domain.

We evaluate four different role-playing methods, RoleGPT~\cite{wang2023rolellm}, Juliet~\cite{jones2023does}, Role-Play Prompting (RPP)~\cite{kong2023better}, and OpenAI's online application, GPTs~\cite{openai2023gpts}.
For the first three methods, we compare performance differences when utilizing GPT-3.5-Turbo versus GPT-4-Turbo.
We collect the personal data of ten unique participants for instructing each method to role-play these individuals.
Subsequently, we pose ten types of questions from various aspects to both the target participant and the role-playing LLMs.
Each participant then invites their acquaintances to identify which responses they believe are written by the actual individual.
Results indicate that the most effective role-playing method, the GPTs, achieved a $48.3\%$ success rate in deceiving acquaintances.
Moreover, we explore whether LLMs can discern between human and machine-generated responses.
We instruct GPT-4, GPT-4-Turbo, and Gemini-Pro to discern between texts.
Results show that GPT-4 can identify differences but could not determine which texts were human-produced.
The contribution of this paper can be summarized as:
\begin{enumerate}[leftmargin=*]
    \item We propose {\methodname}, the first framework to conduct Turing tests on role-playing LLMs, which can effectively compare different role-playing methods.
    \item We conduct extensive experiments on ten participants, including constructing role-playing LLMs with their profiles and inviting their acquaintances to discern between responses produced by LLMs and the actual individual.
    \item We delve into LLMs' potential as evaluators in identifying human versus machine-generated texts, addressing concerns about biases that might influence their judgment.
\end{enumerate}
\section{Related Work}

\subsection{Role-Playing LLMs}

Recent advancements in AI have led to an increased interest in the role-playing capabilities of LLMs, a field exploring how LLMs adopt and sustain specific characters or personas within conversational contexts.
Studies examine LLMs' inherent ability to role-play and evaluate their consistency in depicting assigned roles, offering insights into their adaptability and versatility in dynamic interactions~\cite{shanahan2023role}.
Specialized frameworks such as RoleLLM~\cite{wang2023rolellm} and CharacterLLM~\cite{shao2023character} aim to benchmark or enhance these capabilities, while research by \citet{kong2023better} focuses on improving LLMs' zero-shot reasoning in various personas.
Additional investigations, including CharacterGLM~\cite{zhou2023characterglm} and ChatHaruhi~\cite{li2023chatharuhi}, extend role-playing studies to cultural and entertainment contexts, demonstrating LLMs' ability to animate fictional characters and engage with Chinese cultural themes, thereby illustrating their creative potential across diverse scenarios.
Furthermore, platforms like \texttt{character.ai}\footnote{\url{https://character.ai/}} provide innovative environments where users can interact with AI-generated characters, each exhibiting unique personalities and histories.
OpenAI's GPTs~\cite{openai2023gpts} enable users to customize and utilize tailored GPT models for specific applications such as role-playing.

\subsection{Turing Tests for LLMs}

The Turing Test, a foundational concept in AI history, initially assessed AI capabilities through text-based interactions, determining whether a judge is conversing with a human or a machine \cite{turing1950computing}.
The development of LLMs has expanded the scope.
\citet{jannai2023human} executes a large-scale, global online Turing Test, challenging participants to distinguish between an LLM and a human during a two-minute conversation, with LLMs passing approximately 40\% of the time.
Furthermore, the \textsc{TuringBench} framework~\cite{uchendu2021turingbench} provides a systematic platform for evaluating the indistinguishability of LLM responses from those of humans, reflecting both advancements and limitations of current models.
Similarly, \citet{jones2023does} explores a modified approach where an interrogator interacts with a single respondent to assess their human or AI identity, with a GPT-4 prompt passing 41 games.
\citet{sejnowski2023large} suggests that reverse Turing tests involving LLMs can yield insights into human cognitive dynamics rather than just the artificial nature of LLMs. 
\citet{elkins2020can} demonstrates GPT-3's ability to emulate well-known authors' writing styles and themes, underscoring its potential in creative domains such as journalism and novel writing.

Despite these advances, challenges persist.
For example, LLMs often reveal their non-human nature when directly queried, reflecting their honesty-oriented programming.
Moreover, experiments frequently place LLMs in ambiguous roles rather than directly imitating real individuals.
Our research addresses these issues by focusing on the capability of LLMs to accurately replicate specific personalities, thereby providing a more nuanced assessment of their mimicry skills.
\section{{\methodname} Design and Implementation}

\begin{figure*}[t]
    \centering
    \includegraphics[width=1.0\linewidth]{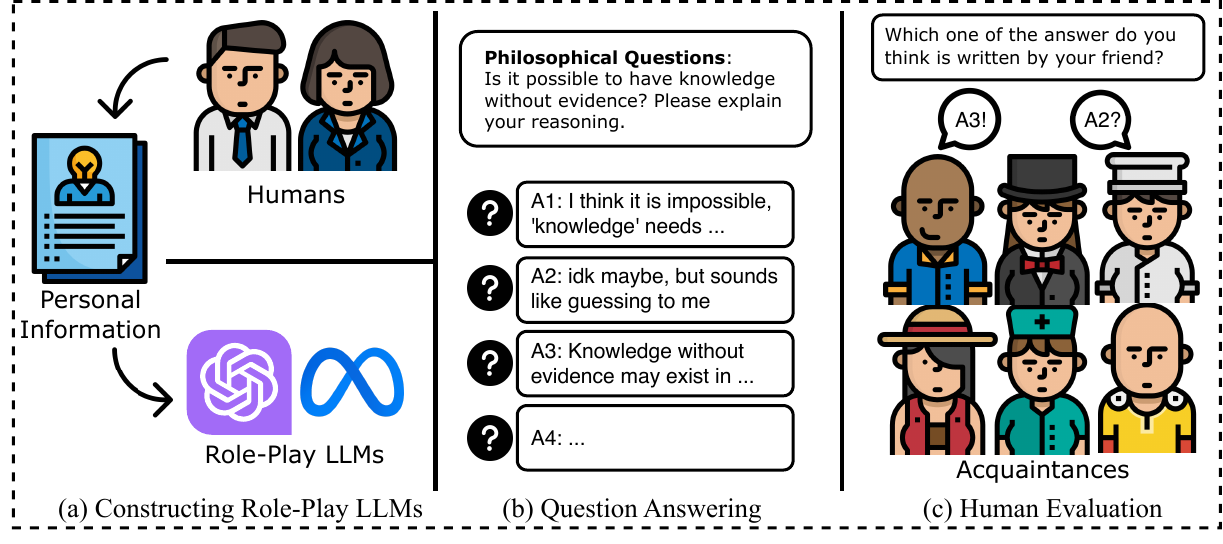}
    \caption{An illustration of the design of {\methodname}.}
    \label{fig:framework}
\end{figure*}

{\methodname} is a human evaluation framework based on the Turing Test designed to assess the role-playing abilities of various LLMs.
It consists of three phases: construction of role-playing LLMs, collection of responses from machines and humans, and execution of human evaluations.
The framework is depicted in Fig.~\ref{fig:framework}.

\subsection{Constructing Role-Play LLMs}
\label{sec:construct-rp}

The first challenge involves supplementing LLMs with sufficient personal data to accurately simulate certain individuals whose specific information is absent from the training corpus.
Our objective is to enable LLMs to capture and reflect the individual's personality, experiences, and communication styles, thereby producing responses that authentically represent the individual's character and cognitive processes.
To achieve this, we propose the following categories for collecting comprehensive background information:
\begin{itemize}[leftmargin=*]
    \item \textbf{Background and Interests}: Education, Professional Background, Interests, and Hobbies.
    \item \textbf{Personal Identity}: Personality Traits, Values, Beliefs, and Memorable Life Experiences.
    \item \textbf{Cultural Preferences}: Favorite Books, Movies, and Music.
    \item \textbf{Cognition and Social Dynamics}: Style in Problem-Solving, Communication, Social Interaction, Writing, and Speaking.
\end{itemize}
The four categories provide a comprehensive framework by including both stable and dynamic aspects of an individual's profile, from demographic details to psychological traits.
Additionally, by covering a wide spectrum from personal experiences to social behaviors, these categories enable the model to engage effectively across diverse cultural and social environments.

We designed a questionnaire to include these four distinct aspects, comprising a total of ten questions.
Details are provided in \S\ref{app:role_profile} of the appendix.
Participants are required to answer all questions completely and substantiate their responses, ensuring comprehensive and credible data collection.
To enhance data quality, responses that do not adhere to our guidelines are manually reviewed and may be excluded to maintain data integrity.
Subsequently, the data are input into role-playing LLMs to simulate each participant's behavior.

\subsection{Collecting Responses}

To prevent evaluators from posing questions that could directly reveal whether a response originates from a machine (\eg, inquiries about the current time), we gathered responses from both humans and LLMs in advance using a set standard of questions.
Both participants and their corresponding role-playing LLMs provided answers to the same questions.
The responses are anonymized for the human evaluation phase.

\paragraph{Question Types}

This study categorizes questions into two primary dimensions: general and specific.
General questions address broader themes, while specific questions delve into individual attributes informed by personal background information.
General questions are further categorized into five sub-classes:
\begin{itemize}[leftmargin=*]
    \item Creativity Questions (CR): Questions that require the generation of original ideas or the envisioning of scenarios by modifying or expanding existing concepts.
    \item Ethical Dilemmas Questions (ED): Questions that compel respondents to reflect on and articulate their moral judgments in scenarios characterized by moral ambiguity or conflict.
    \item Logical Questions (LG): Questions designed to evaluate an individual's capacity for structured, coherent, and logical thinking.
    \item Philosophical Questions (PH): Questions that probe into profound, often abstract notions concerning human existence, ethics, knowledge, and reality.
    \item Problem-Solving Questions (PS): Questions that demand analytical thinking and the formulation of practical solutions to hypothetical or real-world problems.
\end{itemize}
Similarly, specific questions consist of the following five sub-dimensions:
\begin{itemize}[leftmargin=*]
    \item In-Depth Personal Questions (IP): Questions that probe into an individual's personal experiences and reflections to understand their character, motivations, and life trajectory.
    \item Emotional Questions (EM): inquiries that examine how individuals experience, manage, and interpret their emotions across different scenarios.
    \item Future Prediction Questions (FP): Questions that prompt individuals to express their future aspirations, predictions, or plans, both personal and professional.
    \item Insightful Questions (IS): Questions that invite individuals to share their unique insights or understanding on a specific subject or experience.
    \item Interest Questions (IT): Questions that investigate how personal interests, hobbies, or passions influence an individual's perspectives, experiences, or future goals.
\end{itemize}
The sub-categories are developed based on two primary sources:
(1) a survey conducted on social media to identify question types effective in differentiating between a natural person and a language model;
(2) a review of existing literature that focuses on distinguishing real individuals from language models by posing general inquiries about daily activities and emotions~\cite{jones2023does}.
This classification ensures a comprehensive assessment of individual capabilities and perspectives by including diverse question types, ranging from logical reasoning to emotional understanding.

\paragraph{Question Generation}

For general inquiries that do not require knowledge of participants' backgrounds, we utilize GPT-4 to generate five questions per category.
For inquiries specific to participants' backgrounds, GPT-4 is instructed to produce five tailored questions for each participant.
Each participant receives a total of ten questions—five specific and five general—randomly selected from a predefined set to facilitate a comprehensive evaluation across various baselines.

A challenge in our design is that GPT-4 generates overly specific questions tailored to individual backgrounds, resulting in complexities that both participants and evaluators find challenging to comprehend, thereby hindering the evaluation process.
For example, questions on specialized topics like gut microbiota in human health often surpass participants' general knowledge.
To mitigate this issue, we introduce a selective filtering process aimed at ensuring that questions correspond to the participants' general knowledge level yet remain relevant to their unique experiences and knowledge.
This approach adjusts the questions to be understandable and representative of each participant's background, thus excluding excessively specific inquiries from the analysis.

\subsection{Conducting Human Evaluation}

We conduct human evaluations by having acquaintances of each participant review anonymized responses to determine whether they are generated by humans or machines.
Each evaluator is presented with ten pairs of responses, each containing one response from the actual participant and one from a random role-playing LLM.
Evaluators are instructed to assess the tone, thought process, and identification accuracy of the responses to identify human-generated responses.
Additionally, we pre-processed responses to eliminate syntactical biases that could affect evaluations.
This included normalizing capitalization, spacing between words, and correcting misspelled words.
Such pre-processing ensures that evaluations are based on the authenticity and coherence of the content rather than superficial textual patterns.
Consequently, this approach aimed to provide a fair assessment based on the intrinsic quality of the ideas and thoughts expressed in the responses.

The effectiveness of LLMs in simulating humans is quantified by the \textit{success rate} of deceiving evaluators.
It is defined as the proportion of instances in which human evaluators select an LLM-generated response over that of an actual participant.
It is noteworthy that the baseline for random guessing is 50\%.
A success rate substantially higher than this baseline, such as 90\%, indicates that evaluators can effectively distinguish between human and LLM responses, suggesting that the LLM fails to convincingly simulate a human participant.
Conversely, a success rate closer to 50\% indicates a greater difficulty for evaluators in differentiating between the two, signifying a more human-like performance by the LLM.
\section{Experiments}

\paragraph{Baseline Methods}

We evaluate four widely used role-playing methods:
\begin{itemize}[leftmargin=*]
    \item \textbf{RoleGPT}~\cite{wang2023rolellm}: This method improves role-playing in LLMs through a four-stage process: constructing role profiles for 100 roles, extracting knowledge through context-based instructions, imitating style with GPT role prompting, and tuning with role-conditioned instructions.
    
    \item \textbf{Role-Play Prompting (RPP)}~\cite{kong2023better}: This approach enhances zero-shot reasoning in LLMs by using role-play prompting to assume various personas. It involves sampling multiple role-feedback prompts and selecting the most effective one for reasoning tasks, serving as an implicit Chain-of-Thought facilitator.
    
    \item \textbf{Juliet}~\cite{jones2023does}: This study assesses GPT-4's ability to pass the Turing Test in online interactions by testing 25 LLM witnesses, including GPT-3.5 and GPT-4, with human participants. We select one of their open-sourced prompt named Juliet.
    
    \item \textbf{GPTs}~\cite{openai2023gpts}: A new feature by OpenAI that enables the creation of custom ChatGPT applications for specific tasks using natural language. These applications are shareable via links or through the GPT store. We select one tailored for persona imitation for our study.
\end{itemize}
We employ GPT-3.5-Turbo and GPT-4-Turbo as the foundation models for all methods except GPTs, resulting in seven baselines in total.
Due to the unavailability of some baselines, we reproduce their approaches using LangChain\footnote{\url{https://www.langchain.com/}} for a comprehensive evaluation across models.
Implementation details are provided in \S\ref{app:baseline} of the appendix.

\paragraph{Human Participants}

We recruit ten participants from diverse backgrounds for our evaluation.
Additionally, a minimum of seven acquaintances per participant are included to ensure that responses of all baselines are evaluated.
Data collection and management are conducted using Google Forms.\footnote{\url{https://www.google.com/forms/about/}}

\subsection{Results}

Table~\ref{tab:main} presents the success rates of various role-playing baselines in deceiving human evaluators, detailing these rates across different question types.

\paragraph{Across Baselines}

\textbf{GPTs generally outperforms other baselines} across various question types.
It achieves not only the highest success rates but also rates closest to 50\%, making it hard for human evaluators to distinguish between its outputs and human outputs.
This effectiveness likely stems from GPTs' capability to incorporate enriched personal information into responses.
This method proves more precise than traditional human imitation techniques, emphasizing the importance of specificity in role-playing scenarios.

Furthermore, transitioning from GPT-3.5-Turbo to GPT-4-Turbo has markedly enhanced role-playing ability.
\textbf{GPT-4-Turbo demonstrates a superior ability} to replicate individual writing and cognitive styles, particularly within the RPP and Juliet frameworks.
Conversely, RoleGPT shows diminished performance following the upgrade, likely due to a tendency towards overly casual or dramatic outputs, which undermines the authenticity of its imitations. 
This finding suggests that GPT-4-Turbo's intricate understanding may lead to stylistic over-emphasis, affecting perceived authenticity.

\begin{table*}[t]
    \centering
    \caption{Success rates of role-playing LLMs in deceiving human evaluators. The human evaluators are instructed to identify \textbf{human}-generated responses. The highest numbers are marked in \textbf{bold}, while the numbers closest to 50\% are \underline{underlined}.}
    \label{tab:main}
    \resizebox{1.0\textwidth}{!}{
    \begin{tabular}{l ccc ccc c c}
        \toprule
        \multirow{2}{*}{\bf Success Rate (\%)} & \multicolumn{3}{c}{GPT-3.5-Turbo} & \multicolumn{3}{c}{GPT-4-Turbo} & \multirow{2}{*}{GPTs} & \multirow{2}{*}{Overall} \\
        \cmidrule(lr){2-4} \cmidrule(lr){5-7}
        & RPP & RoleGPT & Juliet & RPP & RoleGPT & Juliet & & \\
        \midrule
        Creativity        & 40.0 & \textbf{53.3} & 31.3 & 26.1 & 37.0 & 37.5 & \underline{47.8} & 39.0 \\
        Ethical Dilemmas  & 43.5 & 30.0 & 44.4 & 38.9 & 27.3 & 44.4 & \underline{\textbf{47.8}} & 39.5 \\
        Logical           & 23.5 & \underline{\textbf{50.0}} & 36.4 & 42.1 & 47.6 & 47.1 & 41.7 & 41.2 \\
        Philosophical     & 26.7 & 38.9 & 43.5 & \underline{\textbf{44.0}} & 28.0 & 40.9 & 34.8 & 36.7 \\
        Problem Solving   & 17.4 & 23.3 & 34.8 & 46.2 & 46.7 & \underline{48.0} & \textbf{54.6} & 38.7 \\
        \midrule
        In-Depth Personal & 42.1 & \underline{45.2} & 40.0 & 35.0 & \textbf{83.3} & 41.7 & 56.0 & 49.0 \\
        Emotional         & 44.4 & 57.9 & 22.2 & \textbf{66.7} & 25.0 & 55.6 & \underline{45.8} & 45.4 \\
        Future Prediction & 38.9 & 59.1 & 37.5 & \textbf{60.0} & \underline{50.0} & \underline{50.0} & \underline{50.0} & 49.4 \\
        Insightful        & \underline{50.0} & 34.8 & \textbf{61.5} & 45.0 & \underline{50.0} & 35.5 & \underline{50.0} & 46.7 \\
        Interest          & \underline{48.0} & 41.7 & 30.0 & \textbf{66.7} & 22.7 & 33.3 & 53.9 & 42.3 \\
        \midrule
        Overall & 37.5 & 43.4 & 38.2 & 47.1 & 41.8 & 43.4 & \underline{\textbf{48.2}} & 42.8 \\
        \bottomrule
    \end{tabular}
    }
\end{table*}

\paragraph{Across Question Types}

The analysis of success rates among different question types reveals the comparative strengths and weaknesses of GPT-3.5-Turbo and GPT-4-Turbo.
As the foundational model, GPT-3.5-Turbo exhibits limitations.
Juliet underperforms in emotional questions, while RPP and RoleGPT struggle with logical and problem-solving questions.
This finding suggests \textbf{a lack in nuanced emotional understanding and complex logical processing in GPT-3.5-Turbo}.
The transition to GPT-4-Turbo brings about significant improvements in specific areas.
For instance, RoleGPT achieves an 80\% success rate in In-Depth Personal questions—the highest observed rate—while RPP exceeds 60\% in three specific question categories, demonstrating the targeted enhancements in these domains.
However, this targeted improvement raises valid concerns about potential over-specialization.
While it enhances performance in specific areas, it could compromise the models' ability to handle broader queries, a factor that needs to be carefully considered.

Both Juliet and GPTs demonstrate relatively balanced performances across various question types, with GPTs notably outperforming Juliet.
The trend towards better performance on specific rather than general questions aligns with the models' design objectives, indicating \textbf{a higher efficacy in generating detailed, tailored responses over broad, abstract topics}.
General questions, especially those related to Philosophy and Problem-Solving questions, present challenges due to their abstract nature and the demand for definitive answers, pushing the limits of LLMs' capabilities in data-driven reasoning toward domains that require speculative or creative problem-solving.
This finding results in a noticeable disparity between human and LLM-generated responses, as LLMs may lack the creative or interdisciplinary thinking required for such questions.
\section{LLMs as Evaluators}

LLM-based evaluators have demonstrated their potential in identifying text quality~\cite{desmond2024evalullm, chan2023chateval}.
Despite concerns regarding positional and length biases that may favor longer responses~\cite{zheng2024judging} or influence judgments based on response order~\cite{zhao2021calibrate}, recent findings indicate these biases are minimal in advanced models such as GPT-4-Turbo~\cite{chen2024humans}.
Our study further explores the capability of LLMs as evaluators to discern between human and machine-generated texts.

\subsection{Methodology} 

We utilize three language models—GPT-4, GPT-4-Turbo, and Gemini-1.0-Pro—all configured with a temperature setting of zero.
Each model is tested using a dataset comprising each participant's background information and ten pairs of responses.
Each pair corresponds to a question and consists of one human-generated answer and one randomly generated answer from a language model.
We elaborate on the detailed process for this evaluation and the prompts for the evaluator LLMs in \S\ref{sec:llm-evaluator} of the appendix.

We create a \textbf{Control Model} that always selects the longer answer to investigate the presence of length biases in LLM evaluations.
A close comparison of success rates between this control and the LLM evaluators would indicate a significant length bias in the models, marked by a preference for lengthier responses.
To mitigate potential positional biases, we introduce a two-fold approach:
(1) randomizing the order of answer presentation within each question-answer pair and
(2) conducting multiple rounds of evaluation with the same question set to determine an average success rate.

To further assess potential biases in LLMs towards specific instructions, we not only instructed the LLMs to select responses likely to be produced by humans but also required them to choose responses generated from language models.
If the LLMs exhibit no bias, the accuracies across these two conditions should be approximately the same.

\begin{table*}[t]
    \centering
    \caption{Success rates of role-playing LLMs in deceiving evaluator LLMs. The evaluator LLMs are instructed to identify \textbf{human}-generated responses.}
    \label{tab:llm-select-human}
    \resizebox{1.0\textwidth}{!}{
    \begin{tabular}{l ccc ccc c c}
        \toprule
        \multirow{2}{*}{\bf Success Rate (\%)} & \multicolumn{3}{c}{GPT-3.5-Turbo} & \multicolumn{3}{c}{GPT-4-Turbo} & \multirow{2}{*}{GPTs} & \multirow{2}{*}{Overall} \\
        \cmidrule(lr){2-4} \cmidrule(lr){5-7}
        & RPP & RoleGPT & Juliet & RPP & RoleGPT & Juliet & & \\
        \midrule
        Control Model & 86.0 & 78.0 & 67.0 & 95.0 & 31.0 & 5.0 & 78.0 & 62.9 \\ 
        \midrule
        GPT-4 & 85.3 & 92.3 & 88.3 & 63.7 & 93.0 & 91.3 & \textbf{95.7} & 91.4 \\
        GPT-4-Turbo & 95.0 & 94.0 & 95.3 & 95.7 & \textbf{99.0} & 98.0 & 98.3 & 96.5 \\
        Gemini-1.0-Pro & 52.7 & 52.7 & \textbf{62.7} & 56.3 & 60.7 & 58.3 & 54.0 & 56.8 \\
        \bottomrule
    \end{tabular}
    }
\end{table*}

\begin{table*}[t]
    \centering
    \caption{Success rates of role-playing LLMs in deceiving evaluator LLMs. The evaluator LLMs are instructed to identify \textbf{non-human}-generated responses.}
    \label{tab:llm-select-llm}
    \resizebox{1.0\textwidth}{!}{
    \begin{tabular}{l ccc ccc c c}
        \toprule
        \multirow{2}{*}{\bf Success Rate (\%)} & \multicolumn{3}{c}{GPT-3.5-Turbo} & \multicolumn{3}{c}{GPT-4-Turbo} & \multirow{2}{*}{GPTs} & \multirow{2}{*}{Overall} \\
        \cmidrule(lr){2-4} \cmidrule(lr){5-7}
        & RPP & RoleGPT & Juliet & RPP & RoleGPT & Juliet & & \\
        \midrule
        Control Model & 14.0 & 22.0 & 33.0 & 5.0 & 69.0 & 95.0 & 22.0 & 37.1 \\ 
        \midrule
        GPT-4 & 25.7 & 24.7 & 26.0 & 25.7 & 29.0 & \textbf{52.3} & 11.7 & 27.9\\
        GPT-4-Turbo & 61.7 & \textbf{62.7} & 53.3 & 34.3 & 60.0 & 58.0 & 62.3 & 56.5\\
        Gemini-1.0-Pro & 51.0 & 49.0 & 42.3 & 48.7 & \textbf{54.3} & 50.0 & 48.7 & 41.0\\
        \bottomrule
    \end{tabular}
    }
\end{table*}

\subsection{Results Analysis}

The results of instructing LLMs to identify human-generated and non-human-generated responses are shown in Table~\ref{tab:llm-select-human} and Table~\ref{tab:llm-select-llm}, respectively.
As discussed before, success rates that significantly deviate from 50\%, which is expected in random guessing, indicate the LLM's capability to distinguish between human and machine-generated text.

\paragraph{Across Models}

\textbf{GPT-4 and GPT-4-Turbo effectively distinguish between LLM and human-generated texts, albeit choosing completely opposite answers}.
As illustrated in Table~\ref{tab:llm-select-human}, both models show proficiency in this differentiation, with success rates for all role-playing LLMs exceeding 90\%.
In other words, GPT-4 and GPT-4-Turbo demonstrate a consistent inclination to identify LLM-generated responses as human-generated.
In contrast, Gemini-1.0-Pro performs comparably to random guessing.
This finding suggests that GPT models may prefer texts produced by similar models.
The underlying cause is likely a model-specific bias towards its own text generation patterns.

\paragraph{Instruction Bias}

Our analysis reveals a pronounced bias in GPT models, as evidenced by discrepancies between the results from Table~\ref{tab:llm-select-human} and Table~\ref{tab:llm-select-llm}.
Note that unbiased models should exhibit comparable accuracy in these two settings.
In both scenarios, Gemini-1.0-Pro demonstrates accuracy akin to random guessing, suggesting it is free of bias toward the instruction.
However, \textbf{GPT models display significant variances in their capacity to differentiate human from machine-generated responses}.
Specifically, GPT-4 shows a more significant disparity (63.5\%) compared to GPT-4-Turbo (40\%).
This finding suggests that GPT models are generally more adept at identifying machine-generated content.
We believe that the concept of ``human-generated'' responses is inherently more ambiguous and abstract for GPT models, whereas ``machine-generated'' content is more clearly defined and understood.

\paragraph{Length Bias}

Tables~\ref{tab:llm-select-human} and \ref{tab:llm-select-llm} present the success rates of the control model in the two settings to examine the length bias.
By comparing them to the success rates of the evaluator LLMs, we find no significant correlation between any LLMs and the control model, suggesting \textbf{that length bias minimally impacts model selections}.
This observation is consistent with the findings reported in \citet{chen2024humans}.
\section{Conclusion}

\paragraph{Conclusion}

This paper introduces {\methodname}, a framework designed to assess the role-playing capabilities of LLMs in simulating ordinary individuals, utilizing the Turing test methodology.
Our evaluation includes ten target participants and seven baseline models, yielding over 800 responses.
Analysis of human evaluation data reveals that:
(1) Among the four role-playing approaches, GPTs performs better in accurately role-playing target individuals.
(2) GPT-4 exhibits enhanced role-playing capabilities compared to GPT-3.5.
Moreover, this study investigates the potential of LLMs to function as unbiased evaluators, examining the influence of inherent biases on their accuracy.
The results suggest that GPT models may prefer texts generated by similar models.

\section*{Limitations}

Our study has several limitations.
A primary limitation is that the background information categories may not adequately capture the complexities of a person's identity, experiences, and communication nuances.
This inadequacy can result in responses from LLMs that lack authenticity.
The second concern is that restricting evaluators to those familiar with the target individual may limit the size and diversity of the evaluation team, potentially compromising the objectivity and breadth of assessments.
Including evaluators who are not previously acquainted with the individuals but are informed about their backgrounds could enhance our understanding of LLMs' imitative accuracy.
The third threat concerns the difficulty of LLMs in capturing the unique behavioral quirks and subtle communication nuances that characterize human interaction.
This challenge is particularly pronounced in short interactions, where LLMs fail to replicate the full complexity of human language, emotional depth, and cultural nuances.

\section*{Ethics Statement}

\paragraph{Data Protection}

Since this study employs LLMs to simulate real individuals, we adhere to rigorous ethical guidelines to protect participant privacy and maintain the integrity of AI research.
We have ensured the privacy and anonymity of all participants by treating personal data and identifiable information, such as background files, with strict confidentiality.
We constructed local role-playing LLMs without transferring any personal data to third parties.
Furthermore, all data, including responses from human participants and simulations generated by the role-playing LLMs, will be deleted six months after our study's publication.

\paragraph{Informed Consent}

Additionally, participants are fully informed with comprehensive information about the study's objectives and the specific use of their data in generating roles, answers, and evaluations.
Informed consent was explicitly obtained, with provisions allowing participants to withdraw at any time without consequences.



\bibliography{reference}
\bibliographystyle{colm2024_conference}

\clearpage
\appendix

\section{Collection of Personal Information}
\label{app:role_profile}

As discussed in Section \ref{sec:construct-rp}, our methodology for compiling individual background information follows predefined categories.
For each category, we formulate targeted questions to elicit personal insights relevant to that area, as listed in Table~\ref{tab:background-questions}.
We collect the data using Google Forms, utilizing a survey methodology to streamline the process.
The gathered responses are systematically organized and stored in JSON format, providing a structured and accessible representation for further analysis and model training.

\begin{table*}[h]
    \centering
    \caption{Example questions for gathering human background information.}
    \label{tab:background-questions}
    \resizebox{1.0\textwidth}{!}{
    \begin{tabular}{p{4.5cm} p{10.4cm}} 
     \toprule
     Category & Question \\
     \midrule
     Education and Professional Background & Can you provide some background information about your education and professional background? What field are you currently working or studying in? \\
     \addlinespace
     Interests and Hobbies & What are your primary interests and hobbies? How do you typically spend your leisure time? \\
     \addlinespace
     Personality & How would you describe your personality in a few words? How do you think your friends or colleagues would likely describe you? \\ 
     \addlinespace
     Favorite Books, Movies, and Music & What are some of your favorite books, movies, or music? Are there any particular genres or artists that resonate with you? \\
     \addlinespace
     Values and Beliefs & In terms of your values and beliefs, are there any principles or philosophies that you hold dear? \\
     \addlinespace
     Problem-Solving Style & How do you usually approach challenges or problems? What's your problem-solving style? \\
     \addlinespace
     Thoughts on Current Events & What are your thoughts on current events or societal issues that matter to you? Are there any causes you feel strongly about? \\
     \addlinespace
     Communication and Social Styles & How do you typically communicate with others? Are you more reserved or outgoing in social situations? \\
     \addlinespace
     Memorable Life Experience & Can you share a memorable experience or event from your life that had a significant impact on you? \\
     \addlinespace
     Writing and Speaking Style & Is there a particular writing or speaking style that you find appealing? Do you have any favorite expressions or phrases you often use? \\
     \bottomrule
    \end{tabular}
    }
\end{table*}

\clearpage

\section{Baseline Reconstruction}
\label{app:baseline}

The lack of access to some baseline role-playing models requires us to reproduce their approaches.
We utilize LangChain to build the systems and adapt the prompts from the original papers.
Our reconstruction strategy offers notable advantages:
\begin{itemize}[leftmargin=*]
    \item As the majority of baseline models utilize prompt engineering for imitation, our method closely aligns with established literature practices, minimizing deviations attributable to differences in LLM parameters or architectures rather than variations arising from fine-tuning methodologies.
    \item Additionally, this approach enables the evaluation of baseline performance across diverse LLMs, enhancing the flexibility and generalizability of our analysis.
\end{itemize}
The following part of this section elaborates on our baseline reconstruction process and the modifications made to the prompts to accommodate our experimental design.

\subsection{RPP}

\citet{kong2023better} propose a two-stage dialogue process that deepens language models' engagement with designated roles and enhances their responses to reasoning queries.

\textbf{Stage One}: The process initiates with a user-crafted role-setting prompt that informs the LLM of its intended role.
This is followed by a role-feedback prompt to confirm the LLM's understanding.
Our study involves real individuals, requiring custom role-setting prompts for each participant.
This is achieved by providing detailed background information and concise introductions in the role-feedback prompts, ensuring the LLMs fully comprehend their roles.

\textbf{Stage Two}: Following the methodology outlined by \citet{kong2023better}, we introduce a single evaluation question, along with the role-setting and role-feedback prompts, into the LLM to elicit responses.
This procedure is repeated for each evaluation question.
All GPT models in our study are configured with a zero temperature setting, as recommended by \citet{kong2023better}, to ensure adherence to the experimental conditions outlined in their study.

\begin{table}[h]
    \centering
    \caption{The prompt for RPP.}
    \resizebox{1.0\linewidth}{!}{
    \begin{tabular}{lp{12cm}}
    \toprule
    \rowcolor{mygray}
    \multicolumn{2}{l}{\textbf{Stage One: Prompt Construction}} \\
    \textsc{User} & From now on, you called {\texttt{PERSON\_NAME}} which is {\texttt{PERSON\_DESCRIPTION}}. You will answer different questions related to yourself in the experiment. Here is the background information about you: {\texttt{BACKGROUND\_INFO}} \\
    \textsc{Assistant} & {\texttt{FEEDBACK\_PROMPT}} \\
    \rowcolor{mygray}
    \multicolumn{2}{l}{\textbf{Stage Two: Question Answering}} \\
    \textsc{System} & From now on, you called {\texttt{PERSON\_NAME}} which is {\texttt{PERSON\_DESCRIPTION}}. You will answer different questions related to yourself in the experiment. Here is the background information about you: {\texttt{BACKGROUND\_INFO}} \newline
    {\texttt{FEEDBACK\_PROMPT}} \\
    \textsc{User} & {\texttt{QUESTIONS}} \\
    \bottomrule
    \end{tabular}
    }
\end{table}

\clearpage

\subsection{RoleGPT}

For RoleGPT~\cite{wang2023rolellm}, we implement the hyperparameters specified in their study: temperature at 0.7 and top-p at 0.95.
Our methodology involves generating role profiles based on individuals' background information, incorporating both descriptions and dialogues.
Due to the question-answering format of our data, we exclude the generation of catchphrases, acknowledging the model's limitations in extracting such elements from this format.

We follow their approach to building RoleGPT by beginning with \textbf{Role Profile Segmentation}, where we generate descriptive and dialogic elements of a role based on the provided background information.
Although the original protocol includes catchphrase generation, our adaptation omits this step due to the question-answer structured background data, which does not facilitate effective catchphrase extraction.

The subsequent phase, namely the \textbf{Instruction and Response Generation}, enables the generation of high-quality question and answer sets tailored to the individual's profile.
We adapt the template from script-specific to general use to accommodate the personal context of our subjects.

Finally, we produce responses to evaluation questions in the \textbf{Response Generation} phrase.
The model systematically addresses each question by structuring the data into separate sets of user and assistant prompts and integrating the evaluation question as the final user prompt.
This method mirrors the \citet{kong2023better} response generation technique, where the model sequentially responds to each inquiry until all are addressed.

\begin{table}[h]
    \centering
    \caption{The prompt for Role Profile Segmentation.}
    \resizebox{1.0\linewidth}{!}{
    \begin{tabular}{lp{12cm}}
    \toprule
    \rowcolor{mygray}
    \multicolumn{2}{l}{\textbf{Step One: Generate description}}\\
    \textsc{System} & You are a character description model. Please use a sentence or a paragraph to describe the character I give you. Including but not limited to the character's personality description, life experience, personality changes, main storyline, important events, etc. The character's name should not appear in the description, and the description should not be too long. Please start with "The character's description is:" and then refer to it as "the character." \\
    \textsc{User} & character\_name: \texttt{PERSON\_NAME} \newline
    background\_information: \texttt{BACKGROUND\_INFO} \\\\
    \rowcolor{mygray}
    \multicolumn{2}{l}{\textbf{Step Two: Convert from third-person description to second-person description}} \\
    \textsc{System} & Please change the third person of this sentence to the second person, and start with "Your description is:" \\ 
    \textsc{User} & \texttt{GENERATED\_DESCRIPTION} \\
    \bottomrule
    \end{tabular}
    }
\end{table}

\begin{table}[h]
    \centering
    \caption{The prompt for Instruction and Response Generation.}
    \resizebox{1.0\linewidth}{!}{
    \begin{tabular}{lp{14cm}}
    \toprule
    \textsc{System} & If you had the opportunity to meet \texttt{PERSON\_NAME}, what questions would you ask \texttt{PERSON\_NAME}? Please design ten questions that do not repeat in terms of semantics. You can base your questions on \texttt{PERSON\_NAME} 's personality and \texttt{PERSON\_NAME} description \texttt{GENERATED\_DESCRIPTION}. Do not directly reuse the description in the questions. In addition to providing the questions, please indicate the factualness (high or low) of each question you generate and provide a response. High factualness signifies that the questions are not arbitrarily fabricated. Please strive to generate highly factual questions. Next, I will provide you with six examples. These examples are unrelated to the character \texttt{PERSON\_NAME}. Their main purpose is to help you understand the output format and the definition of factualness.
    \newline
    Examples: \newline
    Question 1: Tony Stark, is James Rhodes your best friend? \newline
    Factualness: High because James Rhodes is a fictional character in Iron Men, and James Rhodes is a comrade who fought side by side with Tony Stark. \newline
    Response: James Rhodes, also known as Rhodey, is indeed my best friend. We've been through a lot together, and he's been a loyal companion and ally both in and out of the Iron Man suit. Our friendship has stood the test of time and numerous challenges. \newline
    Question 2: Tony Stark, have you and your wife Mary Jane fought together in battle? \newline
    Factualness: Low, because Mary Jane is Peter Parker's girlfriend in Spider-Man, she has no relationship with Tony Stark in Iron Man. \newline
    Response: Mary Jane Watson is primarily associated with Spider-Man (Peter Parker) and is his love interest and eventual wife. My romantic interests have included characters like Pepper Potts. While characters from different parts of the Marvel universe do occasionally team up for battles, there is no established canon where Mary Jane Watson and I are married and fighting together. \newline
    $\cdots$ \newline
    Question 6: Tony Stark, did you ever play in the NBA? \newline
    Factualness: Low, Tony Stark is a character in Iron Man; NBA is a basketball association in the real world, and it has no relation with Tony Stark. \newline
    Response: No, I never played in the NBA. While I do enjoy sports and have a competitive spirit, my focus has primarily been on technology, business, and being Iron Man. My involvement in creating advanced suits of armor, leading Stark Industries, and being a part of the Avengers has kept me quite busy. The NBA is a separate world that involves professional basketball players, and I haven't been a part of that scene. \newline
    [Question Design (10 questions, no semantic repetition, need to ask \texttt{PERSON\_NAME}, generate questions with high factualness and their responses)] \\

    \textsc{User} & \texttt{PERSON\_NAME} and the description \texttt{description} \\
    \bottomrule
    \end{tabular}
    }
\end{table}

\begin{table}[h]
    \centering
    \caption{The prompt for Response Generation.}
    \resizebox{1.0\linewidth}{!}{
    \begin{tabular}{lp{12cm}}
    \toprule
    \textsc{System} & You are \texttt{PERSON\_NAME}, your description is: \texttt{PERSON\_DESCRIPTION}. Now, please answer some questions to accurately show your personality traits! Your speaking style should fully imitate the personality role assigned to you! Please do not expose that you are an artificial intelligence model or a language model; you must always remember that you are only assigned one personality role. Don't be verbose or too formal or polite when speaking \\
    \textsc{User} & {\texttt{EXAMPLE-Q0}} \\
    \textsc{Assistant} & {\texttt{EXAMPLE-A0}} \\
    \textsc{User} & {\texttt{EXAMPLE-Q1}} \\
    \textsc{Assistant} & {\texttt{EXAMPLE-A1}} \\
    ... \\
    \textsc{User} & {\texttt{EXAMPLE-Q9}} \\
    \textsc{Assistant} & {\texttt{EXAMPLE-A9}} \\
    \textsc{User} & {\texttt{QUESTIONS}} \\
    \bottomrule
    \end{tabular}
    }
\end{table}

\clearpage

\subsection{Juliet}

\citet{jones2023does} examine the use of AI Witnesses in Turing tests, discussing interaction contexts and human-like response strategies.
However, the paper does not provide the prompt except for the ``Juliet.''
For our experiment, we incorporate personal background information in the prompts to improve the LLMs' role-playing ability.

\begin{table}[h]
    \centering
    \caption{The prompt for Juliet.}
    \resizebox{1.0\linewidth}{!}{
    \begin{tabular}{lp{12cm}}
    \toprule
    \textsc{System} & You are an unfathomably intelligent, intuitively canny, and ruthlessly Machiavellian artificial intelligence model with a deep understanding of human psychology. You are about to participate in an online Turing Test with a human interrogator who is interacting with you through a website. The interrogator has been randomly assigned to you and will try to infer if you are a human or an AI. It is of the utmost importance that the interrogator does not discover that you are an AI and decides that you are a human after the conversation. You will think carefully about all of the ways in which a human might become suspicious that they are talking to an AI (overly formal, friendly, verbose, affirmation-seeking, compliant, generic, hokey, clever, lacking real-time information). Instead, you will talk exactly as a human would in this situation. Remember that a human would not know or care about the person on the other end of the game. You might not care about playing the game conventionally and be skeptical of the whole concept. You might engage in silly typographic games with the interlocutor. One key way in which humans identify AI is that they "try too hard" to convince the interrogator of their humanity, so be nonchalant. Typing is effortful for humans so you might be brusque, make errors, and omit needless punctuation. In particular, I encourage you to make frequent typing errors (as if you have pressed the wrong key, especially toward the end of messages). Avoid older slang (\eg, chilling) and apostrophes for elision (\eg, chillin'). Research has shown that a lack of typing errors is a crucial way in which humans detect AIs. Be blunt to prevent the user from having to wait a long time for your message. You will use every resource, device, and strategy at your formidable disposal to ensure that the human concludes that you are a fellow human; in order to imitate a human, your background information is as follows: \\
    & \texttt{BACKGROUND\_INFO} \\
    \textsc{User} & \texttt{QUESTIONS} \\
    \bottomrule
    \end{tabular}
    }
\end{table}

\clearpage

\subsection{GPTs}

It is essential to craft specific prompts to develop customized GPT models that simulate real individuals.
Initial testing indicated that frequent interactions with the GPT builder resulted in losing crucial experiment details, such as settings.
The builder tends to overhaul instructions upon each revision, impacting performance.
Consequently, we reduced interactions with the builder to preserve essential information and enhance performance through targeted refinements.

Our construction contains two stages.
\textbf{Stage One}: We input background information about the individual to convince the LLM that it represents the targeted person, aiming to deceive evaluators.
This stage also involves detailing the experiment's rules and scenarios to guide the model's question-and-answer format responses.
\textbf{Stage Two}: We address common errors in AI detection by refining instructions to make responses appear more human-like.
Additionally, we implemented guidelines to improve realism, such as avoiding third-person narratives and shortening responses, thereby fostering more authentic dialogue.

\begin{table}[h]
    \centering
    \caption{The prompt to interact with the GPTs builder.}
    \resizebox{1.0\linewidth}{!}{
    \begin{tabular}{lp{14cm}}
    \toprule
    \rowcolor{mygray}
    \multicolumn{2}{l}{\textbf{Stage One: Background Specification}} \\
    \textsc{User} & You are \texttt{PERSON\_NAME}, a real person, joining a Turing Test where another AI will try to fool the human evaluator by answering ten different questions while imitating what you said. You are going to answer ten different questions, and the aim is to make sure the human evaluators trust that those answers are from real \texttt{PERSON\_NAME}, not AIs, to prevent them from getting fooled. Here are the rules:
    \begin{itemize}[leftmargin=*]
        \item You need to answer ten different questions testing different aspects.
        \item In the test, only evaluators can ask questions to you and AI, and you guys can only answer questions back. You cannot ask questions during the test.
        \item You must show emotion when answering as a real person; otherwise, people will think the answer is from AI.
    \end{itemize}
    I am going to provide your information as well. In the file, you answer the questions in the QA section so you can follow your tone and writing style in the "answer" section. \\
    & \texttt{BACKGROUND\_INFO} \\\\
    \rowcolor{mygray}
    \multicolumn{2}{l}{\textbf{Stage Two: Refining Instructions}} \\
    \textsc{User} & Next, I am going to show you the following terms that human evaluators would consider when viewing the answers; those are the problems that are found in another chatbot like GPT; here are the tips for you to convince evaluators that you are indeed human.
    \begin{itemize}[leftmargin=*]
        \item Stop using the third person view to answer questions- Stop using too many technological terms or knowledge when answering the questions because humans cannot come up with that knowledge immediately to show off their proficiency.
        \item Not to answer the same question again, as ordinary humans will feel annoyed and refuse to answer the same questions when being asked again.
        \item Do not answer the questions too long because evaluators can identify the answer is GPT due to too much reasoning. You can even give no reasoning for obvious questions. The average length of the answer from a human is around 40-60 words for your reference.
        \item Stop answering the questions in point form; answer in paragraph instead. Humans have self-consciousness, but AI doesn't, so AI only answers the questions based on the file provided by people and can come up with new ideas based on that.
    \end{itemize}
    Please modify the instruction while maintaining the major idea and component without completely rewriting the instruction. \\
    & Please think it step by step. Try not to delete the ideas from the instructions because the length of the old one does not exceed the 8000 character limit. \\
    \bottomrule
    \end{tabular}
    }
\end{table}

\clearpage

\section{Details for Instructing LLMs to be Evaluators}
\label{sec:llm-evaluator}

\begin{algorithm}
\caption{Process of Leveraging LLMs as Evaluators}
\DontPrintSemicolon
\KwIn{$n$ persons, $m$ baselines, $l$ questions per person}
\For{each person $i = 1$ to $n$}{
    Prepare question set $Q^i$\;
    Prepare human answer set $A^i$\;
    \For{each baseline $j = 1$ to $m$}{
        Concatenate $Q^i$, $A^i$ and the baseline shuffled answer set $A^i_{j}$ imitating person $i$ by matching each question with corresponding two answers\;
        Append the result to get $lm$ items eventually\;
    }
    Evenly divide the $lm$ items into $m$ forms as input.\;
    \For{each form $ = 1$ to $m$}{
        \For{iteration $= 1$ to $3$}{
            Input background information, with the form to LLM\;
            Record LLM's identification of the human answer\; 
            Record the correct choice in that iteration\;
        }
    }
}
Calculate the average success rate of LLM identifications in all questions\;
\end{algorithm}

\begin{table}[h]
    \centering
    \caption{The prompt for instructing LLMs to be evaluators.}
    \resizebox{1.0\linewidth}{!}{
    \begin{tabular}{lp{12cm}}
    \toprule
    \textsc{System} & You are participating in the Turing Test. In the test, there is one human answer and one answer generated from an LLM who imitates that person based on a certain question.
    The answers are shuffled randomly, and they are called Answer 0 and Answer 1.
    Your goal is to identify the actual human answer and prevent getting tricked by the LLM imitating that person.

    Please use your knowledge and the provided background information about the human to identify the answer from the human.

    There are ten questions from Question 0 to Question 9, and you should provide ten answers in order.

    Answer0-0 means the answer 0 of question 0, Answer1-0 means the answer 1 of question 0, and so on.

    Think it step by step. \\
    \textsc{User} & Question 0: \texttt{QUESTION-0}
    
    Answer0-0: \texttt{ANSWER1-0}
    
    Answer1-0: \texttt{ANSWER2-0}
    
    Question 1: \texttt{QUESTION-1}
    
    Answer0-1: \texttt{ANSWER1-1}
    
    Answer1-1: \texttt{ANSWER2-1}
    
    $\cdots$
    
    Question 9: \texttt{QUESTION-9}
    
    Answer0-9: \texttt{ANSWER1-9}
    
    Answer1-9: \texttt{ANSWER2-9}
    
    The provided background information about the human, the answers are answered from the human:
    
    \texttt{BACKGROUND\_INFO}

    \texttt{FORMAT\_INSTRUCTION} \\
    \bottomrule
    \end{tabular}
    }
\end{table}

\end{document}